\def\year{2021}\relax
\documentclass[letterpaper]{article} 
\usepackage{aaai21}  
\usepackage{amsfonts}
\usepackage{times}  
\usepackage{helvet} 
\usepackage{subfigure}
\usepackage{courier}  
\usepackage[hyphens]{url}  
\usepackage{graphicx} 
\usepackage[switch]{lineno} 
\usepackage{natbib}  
\usepackage{caption} 
\usepackage{booktabs}
\usepackage{amsmath}

\newcommand{\etal}{\textit{et al}.}
\newcommand{\ie}{\textit{i}.\textit{e}.}

\title{Fine-grained Identity-Preserving Landmark Synthesis for Face Reenactment}
\author{

    Haichao Zhang\textsuperscript{1,2,3}, Youcheng Ben\textsuperscript{3}, Weixi Zhang\textsuperscript{1}, Tao Chen\textsuperscript{1}, Gang Yu\textsuperscript{3}, Bin Fu\textsuperscript{1,3}\\
}
\affiliations{
    \textsuperscript{1}Fudan University \qquad \qquad \textsuperscript{2}Zhejiang University \qquad \qquad \textsuperscript{3}Tencent\\
    zhanghaichao@zju.edu.cn, \{eugeneben, skicyyu, brianfu\}@tencent.com, \{weixizhang, eetchen\}@fudan.edu.cn
}
\begin{document}
\twocolumn[{%
\renewcommand\twocolumn[1][]{#1}%
\maketitle
\begin{center}
    \centering
    \includegraphics[width=1.0\linewidth]{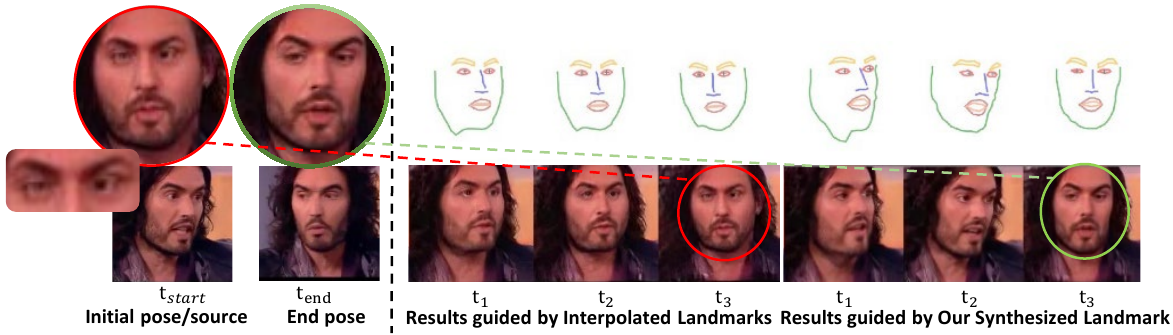}
    \captionof{figure}{Illustration of our method on identity-preserving landmark synthesis for face reenactment. 
    The top row on the right is synthesized landmarks with gradually changing poses from beginning to end using different approaches. The bottom is reenacted face images guided by the above landmarks. 
    As we can see in the red circle, the results guided by interpolated landmarks greatly change the face shape and other facial organs like the unnatural eyes, which leads to an identity preserving problem. In contrast to that, our synthesised landmarks can preserve identity related facial attribute information, and help to reenact better ID-preserved faces.}
\end{center}%
}]


\begin{abstract}
Recent face reenactment works are limited by the coarse reference landmarks, leading to unsatisfactory identity preserving performance due to the distribution gap between the manipulated landmarks and those sampled from a real person. To address this issue, we propose a fine-grained identity-preserving landmark-guided face reenactment approach. The proposed method has two novelties. First, a landmark synthesis network which is designed to generate fine-grained landmark faces with more details. The network refines the manipulated landmarks and generates a smooth and gradually changing face landmark sequence with good identity preserving ability. Second, several novel loss functions including synthesized face identity preserving loss, foreground/background mask loss as well as boundary loss are designed, which aims at synthesizing clear and sharp high-quality faces. Experiments are conducted on our self-collected BeautySelfie and the public VoxCeleb1 datasets. The presented qualitative and quantitative results show that our method can reenact fine-grained higher quality faces with good ID-preserved appearance details, fewer artifacts and clearer boundaries than  state-of-the-art works. Code will be released for reproduction.
\end{abstract}



\begin{figure}[t]
\centering
\begin{minipage}[t]{0.4\textwidth}
\centering
\includegraphics[width=\linewidth]{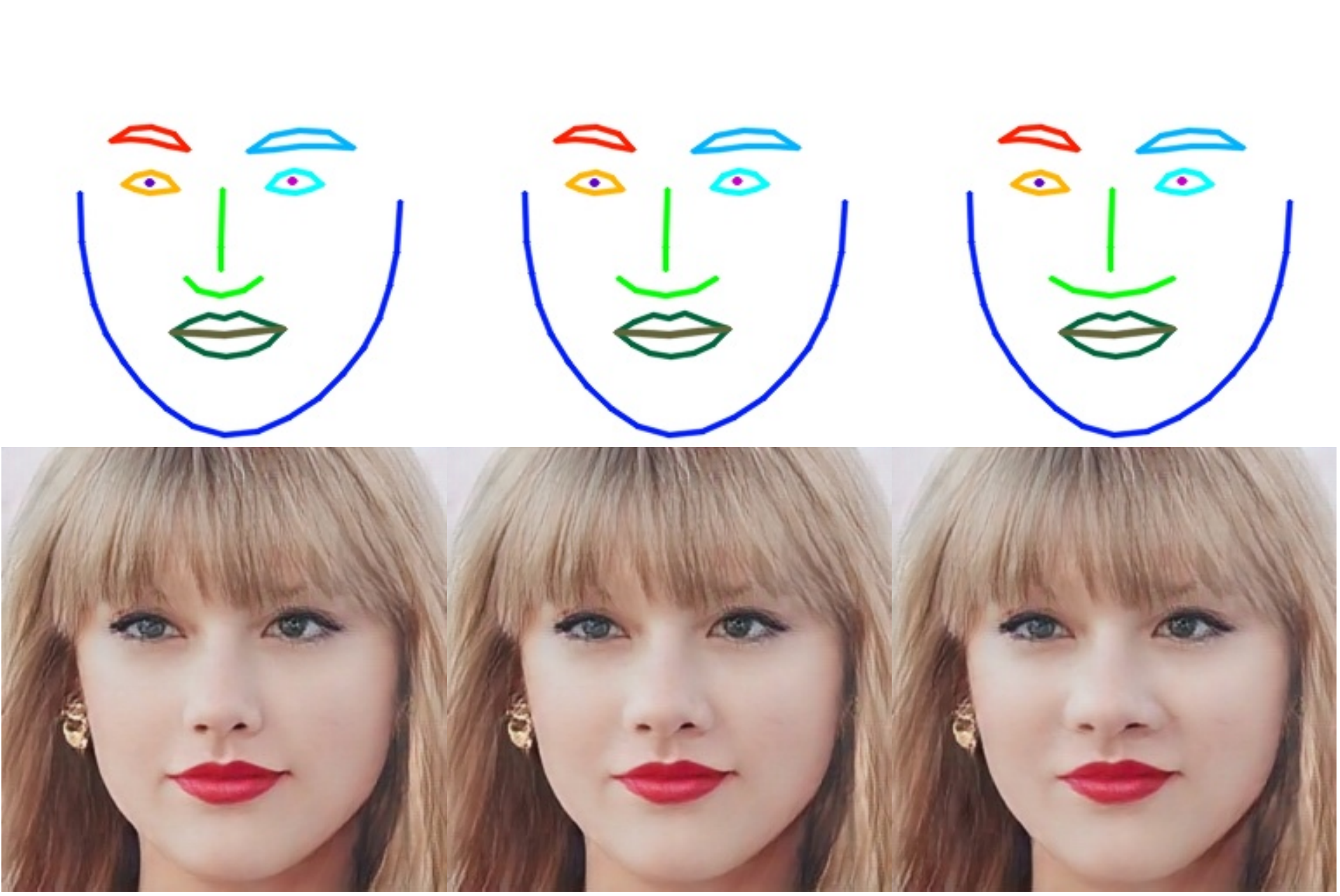}
\caption{nose width (narrow $\to$ wide)} \medskip
\end{minipage}
\caption{Our experiments on manipulating different facial attributes show the existence of the identity gap. 
The identity gap in the manipulated landmarks causes face attribute manipulation to clearly change the identities of reenacted faces.}
\label{fig:landmark}
\end{figure}

\section{Introduction}
\label{Introduction}
Face reenactment is aiming to transfer a reference face landmark including the expression, pose or other details to the target while keeping the appearance and identity of the target unchanged. Take faithful photo-realistic facial reenactment as an example, this technique can find a wide range of applications from video conferencing, film production to augmented reality. Landmark synthesis is widely used as a guided signal of this technique, which has a significant impact on the face reenactment performance. However, due to its geometry structure, the guided face landmark which is supposed to provide emotion information only, often carries the identity information from its provider. This brings difficulty for preserving the identity information of the original face in the face reenactment process. Therefore, an identity-preserving landmark synthesis network for face reenactment is urgently needed.

A number of works related to landmark synthesis and face reenactment have emerged recently. We divide them into three classes. The first class use more diversified inputs than landmark as guided signal for face reenactment, ignoring the identity gap problem. DeepFake~\cite{deepfake} uses an autoencoder network to swap faces of two targets. Tripathy \etal~\cite{tripathy2019icface} use facial attributes as input to guide the source and then synthesize face images with corresponding poses and expressions. Similarly, GANimation~\cite{pumarola2019ganimation} also uses action units(AU)~\cite{au} to synthesize magnitude-controllable expressions. FSGAN~\cite{nirkin2019fsgan} uses segmentation, inpainting and landmark to produce more realistic results. The second class use few-shot training or landmark adaption to convey the emotion, and remove identity information from the target in an explicit or implicit way. Few-Shot~\cite{zakharov2019few},Burkov \cite{burkov2020neural}, FaceSwapNet~\cite{zhang2019faceswapnet} and FReeNet\cite{zhang2020freenet} leverage on landmarks as a tool to restrain the expression of the target. Compared with Few shot-based method~\cite{zakharov2019few}, FaceSwapNet and FReeNet make further progress by optimizing the reference landmarks to make the generated face shape look more like the source. The third class  Face2Face~\cite{thies2016face2face,kim2018deep} generate output faces using 3D models, which can help to suppress the subtle changes of expressions on the synthesized faces.

 Although these recent studies have pushed the research of face reenactment forward a lot, they still suffer from several challenges: 
\textbf{ (1)} Very coarse exemplar landmarks that lose much face content (part) details such as eyeball positions, lip shape or depths, more fine expression and pose types are used, which leads to limited capability to preserve the face appearance and identity information. For example, due to the coarse landmarks used in Few-Shot, FaceSwapNet and FSGAN, these works are difficult to reenact faces with arbitrary and flexible poses and expressions. Even though FSGAN combines landmark, segmentation and inpainting as well as blending proposed in~{\cite{nirkin2019fsgan}} together, the appearance and ID preservation capability of reenacted faces is still not satisfactory.
\textbf{ (2) }Some of the above works, e.g., Few-Shot~\cite{zakharov2019few},Burkov \cite{burkov2020neural},FaceSwapNet~\cite{zhang2019faceswapnet} focus on generating better cross-identity results, but still ignore scenarios that require to synthesize landmarks of the same face identity or maintain identity of the source, which is especially important to application scenarios such as talking heads synthesis~\cite{zakharov2019few}, face animation~\cite{sanyal2019learning} and video synthesis driven by voice~\cite{kumar2017obamanet,zhou2020makeittalk} or text~\cite{fried2019text}. Fig.\ref{fig:landmark} illustrates the identity gap between synthesized landmarks and real landmarks, which makes it difficult to preserve identity information in synthesized faces. 

To address these intractable problems, we propose a fine-grained identity-preserving landmark synthesis approach, which can efficiently generate fine-grained id-preserving landmark sequence for face reenactment according to given two face landmarks. Several novel loss functions including synthesized face identity preserving loss, foreground and background mask loss as well as boundary loss are designed, which aim at synthesizing clear and sharp high-quality faces. Specially, the designed face identity preserving loss consisting of an adversarial training loss and support task, provides supervision for both real/fake generation and the landmark identity prediction, and shows ability to alleviate the identity gap in both explicit and implicit space between two input landmarks. 

Further, the designed foreground and background mask loss makes the generated faces to be more clean and sharp with high quality. Boundary loss improves the results of generated side faces (faces observed from side view) with large poses. Note that the side-face reenactment problem due to large pose changes is not well solved in the above-mentioned works. The large-pose data augmentation and face alignment is leveraged in this work to further improve the large-pose face reenactment results. 

To sum up, our major contributions are listed as follows:
\begin{itemize}
\item  A fine-grained identity-preserved landmark sequence generator is created for alleviating the identity gap problem for face reenactment. As far as we know, this is the fist work focusing on bridging the identity gap in synthesizing landmarks. In addition to landmarks, our model also shows good potential for reducing the identity gap for manipulating faces to a certain degree.
\end{itemize}
\begin{itemize}
\item Several novel strategies including the designed loss functions and large pose data augmentation, aiming at fine-grained exemplar landmark synthesis and face reenactment, are developed, which can help to preserve the landmark and face identity and renders the synthesized results look clean, sharp and high-fidelity. 
\item A largest Asian facial expression dataset, selfie dataset, that contains high-quality video clips of ~23,000 different identities involving most ever seen expressions and poses is created for comprehensive face reenactment and manipulation evaluation.
\end{itemize}

\section{Related Work}
\label{Related Work}
Our work is closely related to conditional face reenactment, identity-preserving synthesis and landmark-guided sequence generation. We discuss main related publications on these research topics as follows. 

\subsection{Conditional Face Reenactment}
\label{GAN-based methods}
Landmark based face reenactment have been widely exploited recently. Many works synthesized faces by introducing landmarks as conditional input in their GAN networks~\cite{goodfellow2014generative}, and the results are really impressive.
FSGAN~\cite{nirkin2019fsgan} presents a better solution with the combination of landmarks, segmentation and inpainting. While it shares certain similarity with our work, it still differs from our work in following aspects.
First, landmarks in FSGAN are mainly used for generating facial expressions of different angles, while our work also leverages landmarks to capture more facial details, such as eyeball movement and lip shapes. Second, the aim of segmentation in FSGAN is to get more concordant blending results of reenactment, while our work uses segmentation mask as restrictions to help generate cleaner and sharper results.

Other works such as GANimation~\cite{pumarola2019ganimation} uses AU~\cite{au} to synthesize magnitude-controllable expressions. Although AU can be easily used to control expressions, it also limits the range of expressions generated in reenactment task. Besides, artifact is another problem of this method.


\subsection{Identity Preserving Landmark Synthesis}
\label{identity-preserving}
To preserve identity information of source faces during generation, previous works such as Few-Shot~\cite{zakharov2019few}, FaceSwapNet~\cite{zhang2019faceswapnet} guarantees identity by imposing restrictions to synthesized landmarks. In terms of many-to-many face reenactment, recent works~\cite{huang2020learning,zhang2020freenet,burkov2020neural} have explored effects of landmarks on the identity preserving of synthesized faces, and the goal is to achieve identity-invariant face reenactment. 

However, most recent works focus on generating better cross-identity results, with less attention paid to the quality of manipulated landmarks for the same face identity, which is actually crucial to application scenarios such as talking heads synthesis~\cite{zakharov2019few}, face animation~\cite{sanyal2019learning} and video synthesis driven by voice~\cite{kumar2017obamanet,zhou2020makeittalk} or text~\cite{fried2019text}. The gap between manipulated and real landmarks makes it difficult to preserve source identity information in reenacted faces, and in this work we aim to reduce this gap.

\begin{figure*}
\begin{center}
\includegraphics[width=0.7\linewidth]{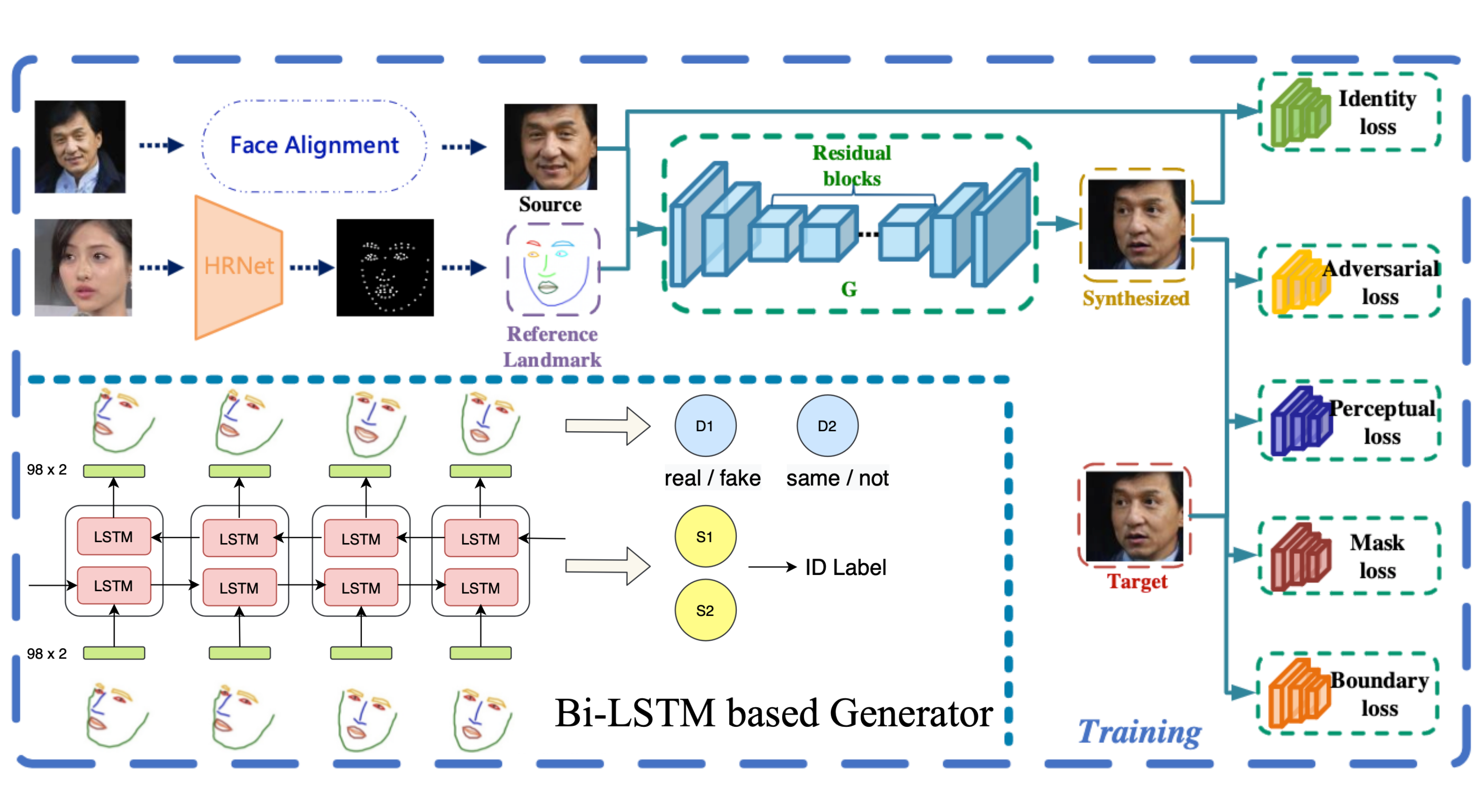}
\end{center}
   \caption{Framework of our proposed fine-grained landmark guided face reenactment network. 
   The left bottom upper branch is Framework of our proposed identity-preserved
landmark sequence generator. Our Bi-LSTM based generator G1 takes two
facial landmarks to predict six medium landmark states of
the inputs. The D1, D2 are two discriminators providing adversarial
losses, while S1 and S2 are two networks providing
the support task losses. The D1 and D2 predict the realness
score of the output landmarks and the probability that the input
and output belongs to the same id, respectively. The S1,
S2 try to predict ID label from the hidden state feature and
output landmarks of G1, respectively.
   The upper branch is the training stage and the lower branch is the inference stage. In the training stage, an input face goes through face alignment and forwarded to the generator $G$ together with an exemplar reference landmark created by a pretrained network. Several newly designed losses: arc loss, mask loss and boundary loss coupled with perceptual and adversarial loss are used as training objectives.}
\label{fig:framework}
\end{figure*}

\section{The Proposed Approach}
\label{sec:model}

In this section, we will present the details of the proposed landmark synthesis guided face reenactment approach. It can be divided into two stages. In the first stage, our main objective is to predict a facial landmark sequence which keeps the identity related information unchanged and approaches the real temporal sequence distribution. In the second stage, our main objective is to transfer pose and expression from one face to another. Given a source face image $F_s$ which contains the identity information and a target face image $F_t$ which provides the pose and expression guidance, our model is supposed to synthesize a new face which retains the appearance of $F_s$ and the pose and expression of $F_t$.  


The overall architecture of our approach is described in Section~\ref{sec:model_0}. Then we describe the identity-preserved landmark sequence generator in section~\ref{sec:model_1}.
Training objectives and data augmentation strategy are detailed in Section~\ref{sec:model_3} and Section~\ref{sec:model_4} respectively.

\subsection{Identity-preserved Landmark Sequence Generator.}
\label{sec:model_1}
Figure~\ref{fig:framework} illustrates the framework of our proposed identity-preserved landmark sequence generator (LSG) . 
We adopted a Bi-LSTM auto-encoder $G_l$ as the backbone network for our identity-preserved landmark generation.
The landmark generator takes two sets of face keypoints $p_{s1}, p_{sk}$  from two face images by HRNet\cite{SunXLW19} as input. We use $[p_{s1}, p_{sk}]$ to denote the input, and the $[p_1,p_2,...,p_K], (K\in N)$ to denote the output,where $k$ is number of generated landmark frames. During training process, we use realness discriminator $D_1$ and identity consistence discriminator $D_2$ to provide adversarial loss, and use two fully-connected nets to predict the id label from each output landmark $p_k, k\in K$ and hidden state features of $G_1$, respectively.

\textbf{ Forward  Process.\quad}
 Firstly, the $[p_{s1}, p_{sk}]$  are upsampled to $K$ frames, then $G_l$ takes them as the input and predicts shifts for each key point. The forward process can be denoted as:
\begin{equation}
\begin{split}
[ p_1,...,p_K ] =  &\Delta shift  + f_{up}[p_{s1},p_{sk}] \\
=  &G_l(p_{s1}, p_{s1},..,p_{sk}) + [p_{s1},,..,p_{sk}]
\end{split}
\end{equation}
in which $f_{up}$ denotes the upsampling process. The overall loss function can be denoted as:
\begin{equation}
\mathbf{L}_{LSG} = \lambda_1 \mathcal{L}_{D_1}+ \lambda_2 \mathcal{L}_{D_2}+ \lambda_3 \mathcal{L}_{S_1} + \lambda_4 \mathcal{L}_{S_2}
\end{equation}
where the $\lambda_{i},(i=1,2,3,4)$ denotes the weight of each loss function.

\textbf{Support task Loss}
Huang \cite{huang2020learning} uses an ID classifier to remove the identity feature in landmark latent space. However, we find that such way of using ID classifier could encourage our hidden state to take the identity information from the input landmark. We employ $S_2$ to predict the ID label of input face key points $p_{s1}, p_{sk}$ , and use the cross entropy loss of $S_2$ to force our LSG to learn identity information as much as possible. So the $\mathcal{L}_{S_2}$ can be denoted as:
 \begin{equation}
 \mathcal{L}_{S_2} = \mathcal{L}_{CE}(c_{p},c_{S_2})
 \end{equation}here, $c_p$ and $c_{S_2}$ represents identity class label of input $p_{s1}, p_{sk}$ .
 In the explicit way, $S_1$  predicts the ID label and provide cross-entropy loss from generated landmarks.
 \begin{equation}
 \mathcal{L}_{S_2} = \mathcal{L}_{CE}(c_{p},c_{S_1}(p_k)),k\in[1,K]
 \end{equation}
 
\textbf{Adversarial Loss}
We employee two discriminator $D_1$, $D_2$ to provide adversarial losses. The $D_1$ takes one landmark each time, to predict whether the landmark is synthesized by $G_1$ or sampled from real world. The $D_2$ takes two landmarks and predict the possibility that they belongs to the same identity.

\subsection{Fine-grained landmark-guided face reenactment }
\label{sec:model_0}
Figure~\ref{fig:framework}  illustrates the pipeline of our proposed approach for fine-grained landmark guided face reenactment. We adopt Pix2pixHD~\cite{wang2018pix2pixHD} as the backbone network for face image generation. Besides the source input image $F_s$, our model also takes as input the sketch map $p_t$ of a target face image $F_t$, which is obtained based on $98$ facial landmarks predicted by HRNet~\cite{SunXLW19}. Therefore, the input to our model is a concatenated volume $x = [F_s, p_t], x\in \mathcal{R}^{H\times W\times (3+3)}$. The output of our model is a volume $y\in\mathcal{R}^{H\times W\times3}$ which refers to a synthesized face image. During the inference stage, the ID-preserving landmark generator is used to produce more fine-grained landmarks with different expressions and poses, which are used to reenact more faces with fine-grained expression and pose details.

\textbf{Training Objectives}
\label{sec:model_3}
For training the face reenactment network, the adversarial loss~\cite{goodfellow2014generative,wang2018pix2pixHD} and perceptual loss~\cite{johnson2016perceptual} are adopted to generate faces with high fidelity. Besides, we also propose novel boundary loss and mask loss to ensure fewer artifacts and sharper details on the boundary and surface of synthesized faces respectively. Furthermore, identity loss is explored which effectively retains the appearance information of the source face $F_s$.

\subsubsection{Adversarial Loss}
To make our synthesized faces more realistic, we employ an adversarial objective to push the distribution of generated images as close as possible to that of real images. Similar to Pix2pixHD~\cite{wang2018pix2pixHD}, we adopt a multi-scale discriminator consisting of multiple discriminators, $D_1, D_2, \dots, D_n$, with each operating on a different image resolution. For a generator $G$ and a multi-scale discriminator $D$, our adversarial loss is defined as:
\begin{equation}
    \mathcal{L}_{adv}(G, D) = \min_{G}\max_{D_1, \dots, D_n}\sum_{i=1}^{n}\mathcal{L}_{GAN}(G, D_i),
\end{equation}


\subsubsection{Identity Loss}
To preserve the identity information of source face $F_s$, we propose an identity consistency loss which regularizes the synthesized face to be close to $F_s$ in the ArcFace~\cite{deng2019arcface} feature space, which enables the generator to preserve the face identity effectively. The identity consistency loss is defined as:
\begin{equation}
    \mathcal{L}_{ID}(G) = \mathbb{E}_{x}[d_{cos}(\theta(G(x)), \theta(x))],
\end{equation}
Here, the $\theta$ is a pre-trained ArcFace~\cite{deng2019arcface} network. 

\subsubsection{Mask Loss}
Generally, the synthesis of foreground facial details is more important than that of background. Therefore, we attempt to segment the facial area of the ground truth to obtain a foreground mask. Leveraging on that mask, we introduce pixel-level $L1$ loss between the generated images and real images in order to reduce artifacts and make the synthesized faces more clear. The mask loss is defined as:
\begin{equation}
    \mathcal{L}_{MASK}(G) = \mathbb{E}_{x,y}[||G(x)\odot M, y\odot M||_1],
\end{equation}
where $M$ is a foreground attention mask parsered by a pre-trained BiSeNet~\cite{yu2018bisenet}.

\subsubsection{Boundary Loss}
As described before, we adopt HRNet~\cite{SunXLW19} to predict facial landmarks of both synthesized and real images. By aligning landmarks on the boundary of real and synthesized faces as close as possible, we expect our model to generate images with fewer artifacts on the boundary of face area. The boundary loss is defined as:
\begin{equation}
    \mathcal{L}_{BOUND}(G) = \mathbb{E}_{x,y}[d_{smoothL1}(\eta(G(x)),\eta(y))],
\end{equation}
where $\eta$ is a pre-trained HRNet~\cite{SunXLW19} on WFLW~\cite{wu2018look}, and smooth L1 distance is implemented as~\cite{girshick2015fast}.

\begin{figure}[t]
\begin{center}
\includegraphics[width=1.0\linewidth]{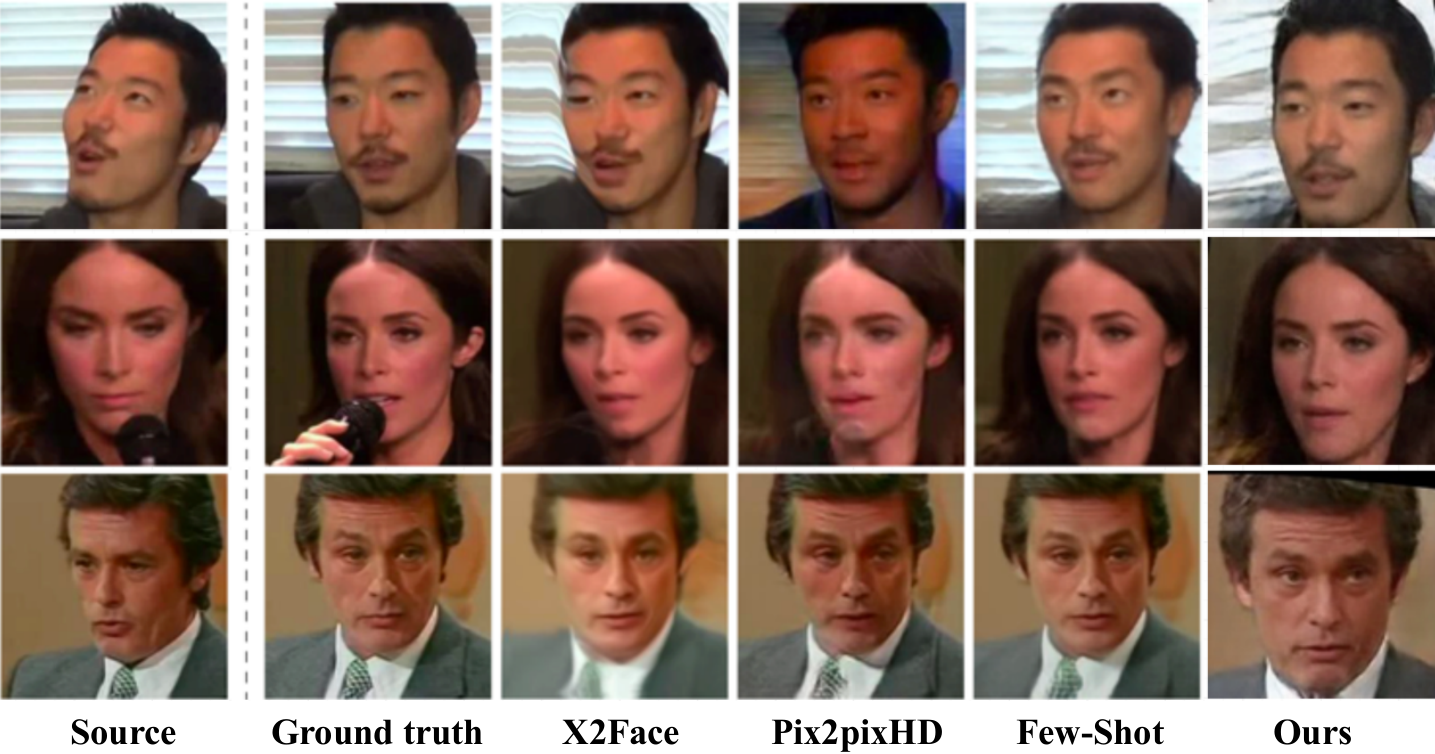}
\end{center}
\caption{Qualitative comparison of different approaches on the VoxCeleb1 dataset. One-shot inference on an identity not seen during training is shown in the \textbf{source} column. Next columns show \textbf{ground truth} image, taken from the rest part of the video sequence, and the synthesized results of different approaches. Note that our result is slightly different from others due to customed face alignment applied to the whole dataset. }
\label{fig:qualitative}
\end{figure}

\subsection{Pose-Aware Data Augmentation}
\label{sec:model_4}

To deal with large pose transfer cases or side-view face generation problem, we develop a novel pose-aware data augmentation strategy, which gradually increases the difficulty of pose transfer during training. In particular, for video frames of each identity in the dataset, we first gather statistics of maximum pose variation range in terms of Euler angles(\ie, yaw and pitch). Then we set an initial threshold based on the computed pose statistics to determine whether to sample from specific identity or not. If the identity pose angle is above the threshold, it will be sampled for training. After every tens of training epochs, we gradually increase the threshold to allow for sampling more difficult pose transfer cases.

There are two advantages by adopting this augmentation policy. One is that our model is capable of learning pose transfer more smoothly along with gradually increased pose sampling threshold. The other is that under this sampling policy, our model is likely to observe more images with side-view faces, which can synthesize better face results with fewer artifacts.
\section{Experiments}
\label{sec:exp}
To showcase the effectiveness of our proposed approach, we perform comprehensive studies on two video datasets involving facial movements and compare our model with several state-of-the-art approaches.

\subsection{Datasets}
We evaluate our model on both public VoxCeleb1~\cite{nagrani2017voxceleb} dataset and our self-collected BeautySelfie dataset. 
BeautySelfie contains $37k$ face videos of beautified selfies, 
each corresponding to a particular identity, of which we use $10k$ for training which are of good quality. 
For fair and fast comparison, we only conduct ablation studies on this private dataset. VoxCeleb1~\cite{nagrani2017voxceleb} (256p videos at 1fps) is a famous dataset consisting of talking head videos, which contains over $100,000$ utterances for $1,251$ celebrities. 
We follow~\cite{wiles2018x2face,tripathy2019icface} to perform train/test splits on this dataset. Since VoxCeleb1 is a public benchmark for face reenactment, we perform both qualitative and quantitative comparisons with previous works on this dataset.

\subsection{Implementation details}
For both datasets, we preprocess raw video frames by first detecting faces using a pre-trained face detector. We filter out cropped faces which are blurry or have lower resolution than $128\times128$ pixels. Then we align faces based on facial landmarks predicted by HRNet~\cite{SunXLW19} under least square constraints. We fill missing pixels with zeros which may result from rotation operations during the alignment. Finally, faces are rescaled to $256\times256$ pixels in all experimental settings. 


\begin{figure}[t]
\begin{center}
\includegraphics[width=1.0\linewidth]{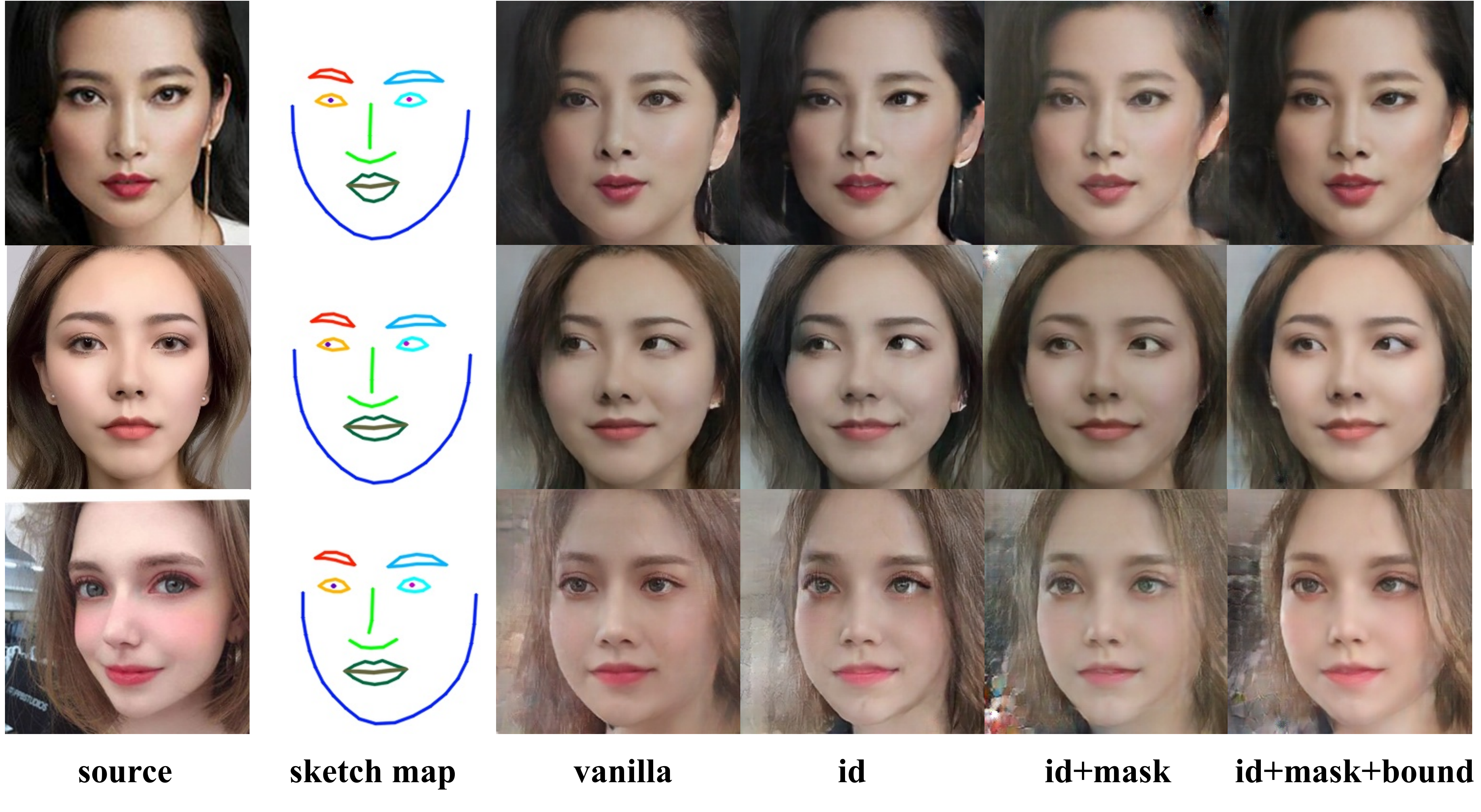}
\end{center}
\caption{ Ablation study of the effectiveness of different loss functions on the BeautySelfie dataset. One-shot inference on an identity not seen during training is shown in the \textbf{source} column. \textbf{Sketch map} corresponds to facial landmarks predicted by HRNet on a target face from the same identity. Result of the model trained with only adversarial and perceptual loss is shown in the \textbf{vanilla} column. Results of models trained with additional identity loss, mask loss and boundary loss are shown in the next three columns respectively. }
\label{fig:ab1}
\end{figure}

In each training iteration, we sample source faces $F_s$ and target faces $F_t$ from the same video sequences to construct training pairs for face reenactment. We sample $10$ pairs of training examples iteratively for each identity in the dataset during this process. The proposed model is trained from scratch, where the weights are initialized randomly using a normal distribution. We use Adam~\cite{kingma2014adam} optimization ($\beta_1 = 0.5, \beta_2 = 0.999$) for both generator and discriminator with an initial learning rate of $0.0002$. We train our network for a total of $45$ epochs and reduce the learning rate linearly to zero after $30$ epochs. 

\textit{Evaluation metrics.}~~For quantitative comparison, we evaluate two indexes, \ie the fidelity and the semantics of the synthesized results. For the evaluation of reality and diversity, we adopt the widely used \textbf{Fr\'echet Inception Distance}~\cite{heusel2017gans} metric. In terms of the semantics, we use \textbf{SSIM}~\cite{wang2004image} to measure the low-level similary between synthesized faces and expected real faces, and \textbf{CSIM}~\cite{zakharov2019few} to calculate cosine similarity between embedding vectors of the state-of-the-art face recognition network~\cite{deng2019arcface} for measuring identity mismatch. These two metrics to some extent reveal how successfully we tranfer pose and expression from a source face $F_s$ to a target face $F_t$ while retaining the appearance information of $F_s$.

\subsubsection{Qualitative Landmark Synthesis Performance}
\begin{figure}[h]
\begin{center}
\includegraphics[width=1\linewidth]{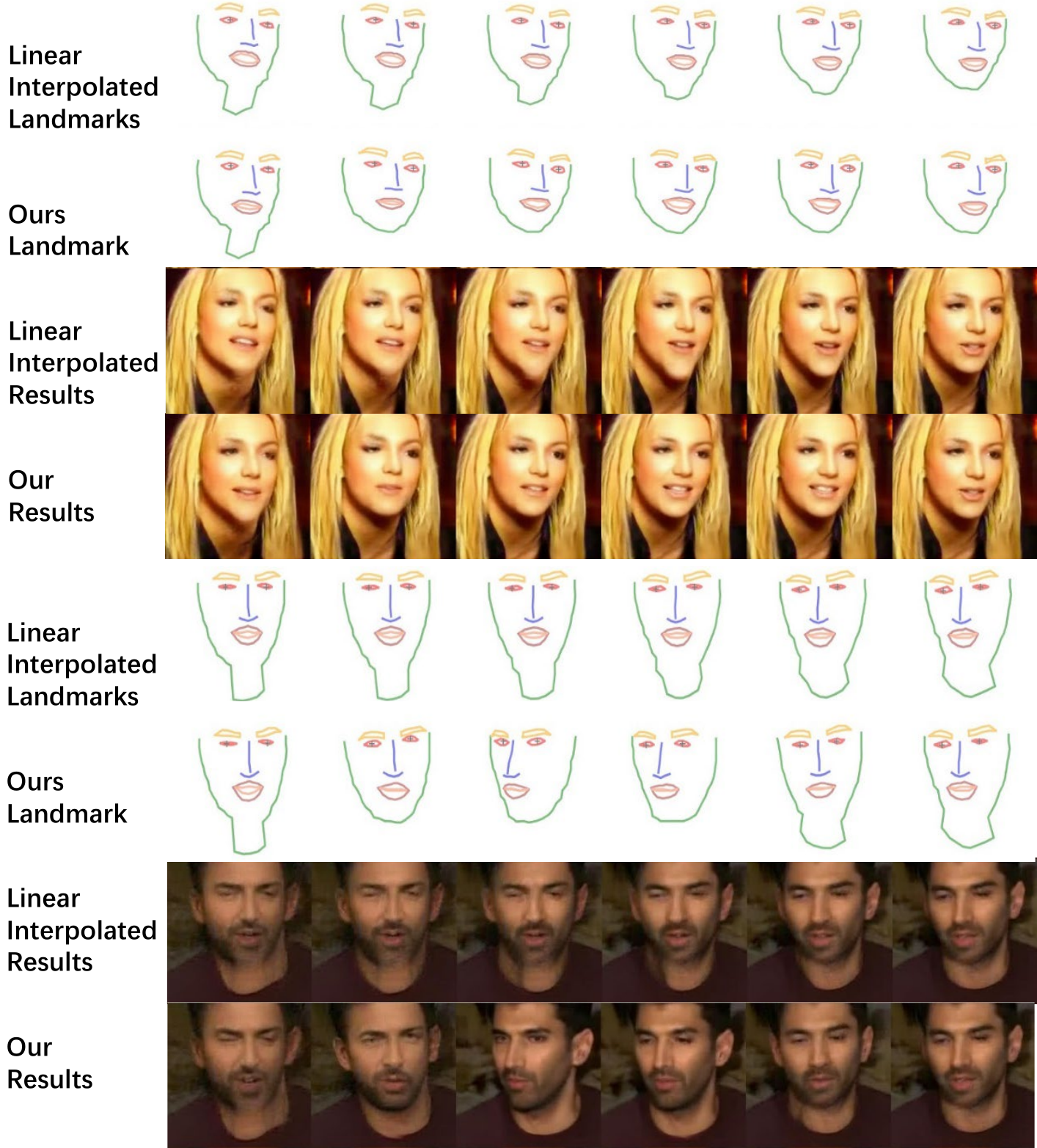}
\end{center}
\caption{ The refinement phenomenon appeared in our landmark generator. With two input noisy landmarks, the landmark sequences was generated by linear interpolation and our approaches. Due to the linear interpolation calculation, the left and right landmark in the Linear Interpolation Landmarks are the same as the two landmarks. Our Landmarks generator show the ability to refine the noise in the input landmarks, which prevents the face reenactment results with less artifacts, and in this situation, better preserving the identities.}
\label{fig: refine}
\end{figure}

Fig.\ref{fig: refine} shows the refinement phenomenon in our landmark synthesis model. The two landmark generation approaches take two noisy input landmarks, and predict six landmarks. The synthesised landmarks was then send them into a pix2pixHD generator to reenact the synthesised face. 
Our landmarks generator show the ability to refine and generate natural landmarks with the noisy input, in the meanwhile, the linear interpolation has nothing to do with it. 
This experiment also shows the robustness and potential refinement ability of our approaches.

\subsubsection{Qualitative Face Reenactment Performance}
Figure~\ref{fig:qualitative} compares synthesized one-shot results of several approaches on the VoxCeleb1~\cite{nagrani2017voxceleb} dataset, where we quote results of X2Face, Pix2pixHD and Few-Shot directly from~\cite{zakharov2019few}. According to~\cite{zakharov2019few}, X2Face is used with pre-trained weights provided by the original authors, while both Pix2pixHD and Few-Shot are pre-trained from scratch on the \textit{whole} dataset. Note that these approaches are also fine-tined \textit{additionally} on the few-shot set. In comparison, our model is learned only on the training set.

Compared to previous approaches, our model is capable of generating more realistic human faces with fewer artifacts. Further, the reenacted faces of our approach meets the target better from different aspects, i.e. eyeball positions, lip or mouth shapes and face poses. More importantly, we observe more fine-grained details in our synthesized faces that are consistent with the source faces. This shows that our model has better capability to retain appearance information of the source faces and transfer target expression or pose information for reenacting results.

\begin{figure}[h]
\begin{center}
\includegraphics[width=0.8\linewidth]{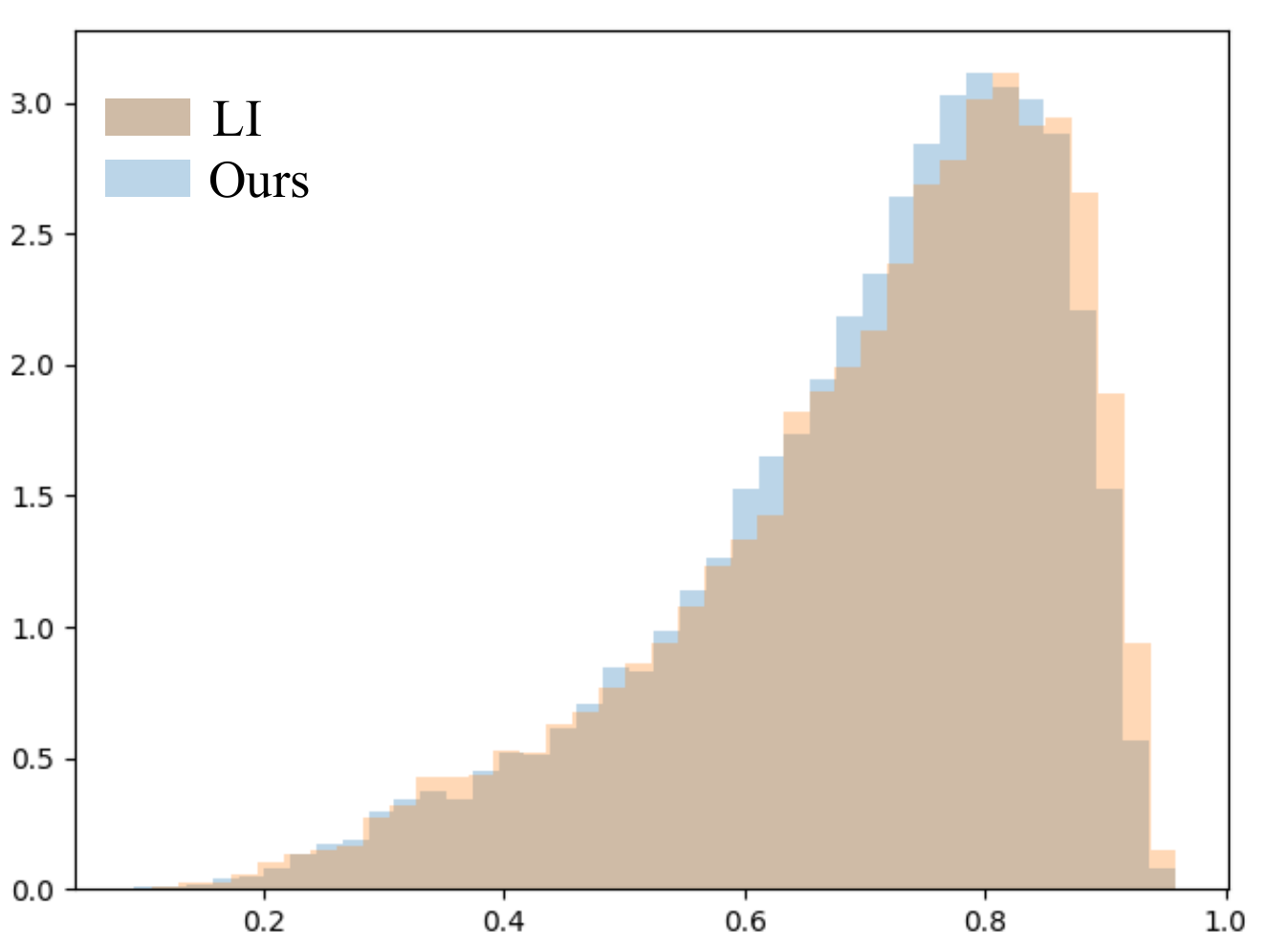}
\end{center}
\caption{ Histogram of CSIM distribution quantitative comparison of our Landmark Synthesis Performance approaches in  Table\ref{tab:Landmark} , where horizontal axis and vertical axis denote the CSIM score and relative numbers of samples perform this score.}
\label{fig:histogram}
\end{figure}

\subsection{Quantitative Results}
\subsubsection{Quantitative Results on Landmark Synthesis Performance}
\label{qua_res}
Table \ref{tab:Landmark} and Fig.\ref{fig:histogram} compare the face reenactment result with different approaches of synthesized landmark, with the metric cosine similarity (CSIM)\cite{zakharov2019few} to mearsure the identity similarity. Due to few works focus on the identity-preserving landmark synthesis, we use the linear interpolation (LI)  result to simulate a smooth and gradually changing between two input landmarks as our baseline.  

Our synthesized landmark has better performance on CSIM score, which indicates our synthesized landmark preserve better identities than LI baseline. The histogram shows the distribution of our method has more ares in the score range 0.80-0.97, while the different area of LI nearly with the same size is in lower score range.  To sum up, our ID-preserving model has better identity-preserving ability than the baseline.

\begin{table}
\begin{center}
\begin{tabular}{l c}
\toprule
Method & CSIM$\uparrow$ \\
\hline 
\multicolumn{2}{c}{VoxCeleb1} \\
\hline
LI & $0.70$ \\
Ours & $0.72$\\
\bottomrule
\end{tabular}
\end{center}
\caption{ Quantitative comparison of our Landmark Synthesis Performance approaches with LI baseline.}
\label{tab:Landmark}
\end{table}

\subsubsection{Quantitative Results on Face Reenactment Performance.}
\begin{table}
\begin{center}
\begin{tabular}{l c c c}
\toprule
Method & FID$\downarrow$ & SSIM$\uparrow$ & CSIM$\uparrow$ \\
\hline 
\multicolumn{4}{c}{VoxCeleb1} \\
\hline
X2Face & $45.8$ & $0.68$ & $0.16$ \\
Pix2pixHD & $\mathbf{42.7}$ & $0.56$ & $0.09$ \\
Few-Shot & $43.0$ & $0.67$ & $0.15$ \\
\hline
Ours & $43.9$ & $\mathbf{0.79}$ & $\mathbf{0.18}$\\
\bottomrule
\end{tabular}
\end{center}
\caption{ Quantitative comparison of several approaches under \textit{one-shot} experimental setting with three different metrics on the VoxCeleb1 dataset. The baselines here are X2Face~\cite{wiles2018x2face}, Pix2pixHD~\cite{wang2018pix2pixHD} and Few-Shot~\cite{zakharov2019few}.}
\label{tab:qualitative}
\end{table}

Table~\ref{tab:qualitative} compares synthesized one-shot results of several approaches on the VoxCeleb1 dataset with three different metrics, where we quote results of X2Face, Pix2pixHD and Few-Shot from~\cite{zakharov2019few} likewise in section~\ref{qua_res}.

Our synthesized results have similar fidelity compared to other approaches in terms of the FID metric. For the other two metrics, SSIM and CSIM, significant improvements are observed. In particular, our model increases the SSIM score of X2Face by 11\%, and CSIM score of X2Face by as high as 2\%, which reveals that our model is superior to other approaches in transferring the semantics of pose and expression while retaining the identity information.

\subsection{Ablation Study}

To systematically evaluate the proposed method, we perform comprehensive ablation studies using different combinations of the proposed loss functions (\ie identity loss, mask loss and boundary loss) and pose-aware data augmentation policy.

\textbf{Effect of the proposed loss functions.}~~Figure~\ref{fig:ab1} demonstrates synthesized results of our model under different groups of loss functions on the BeautySelfie dataset. Results in the vanilla column provide a strong baseline for comparison. We sequentially compare the last three columns with the vanilla column to figure out the effect of each proposed loss function.

Compared to synthesized faces in column $3$, we observe better synthesis results in column $4$ with nose, mouth, eyes and other fine-grained details more consistent with source, which reveals that the identity preserving loss effectively retains appearance and part information of source images. As a comparison between synthesized faces in column $4$ and $5$, it can be seen that the mask loss helps eliminate artifacts on the surface of synthesized faces. As for results displayed in column $6$, we find boundary loss enables our model to generate faces with better boundary lines and fewer artifacts around.

\begin{figure}[t]
    \centering 
    \subfigure[without pose-aware data argumentation]{
    \label{fig:a}
    \includegraphics[width=0.45\linewidth]{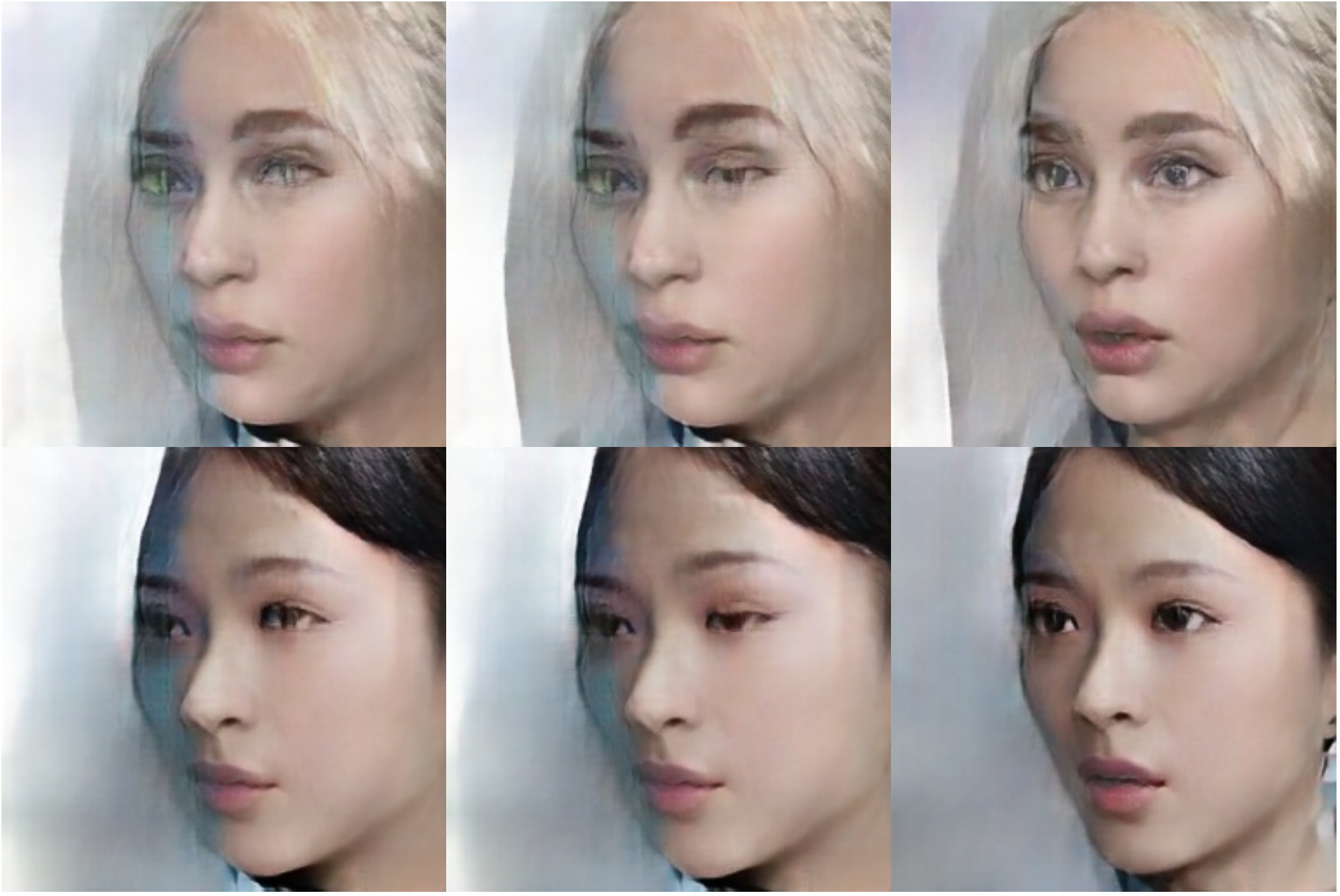}
    }
    \subfigure[with pose-aware data argumentation]{
    \label{fig:b}
    \includegraphics[width=0.45\linewidth]{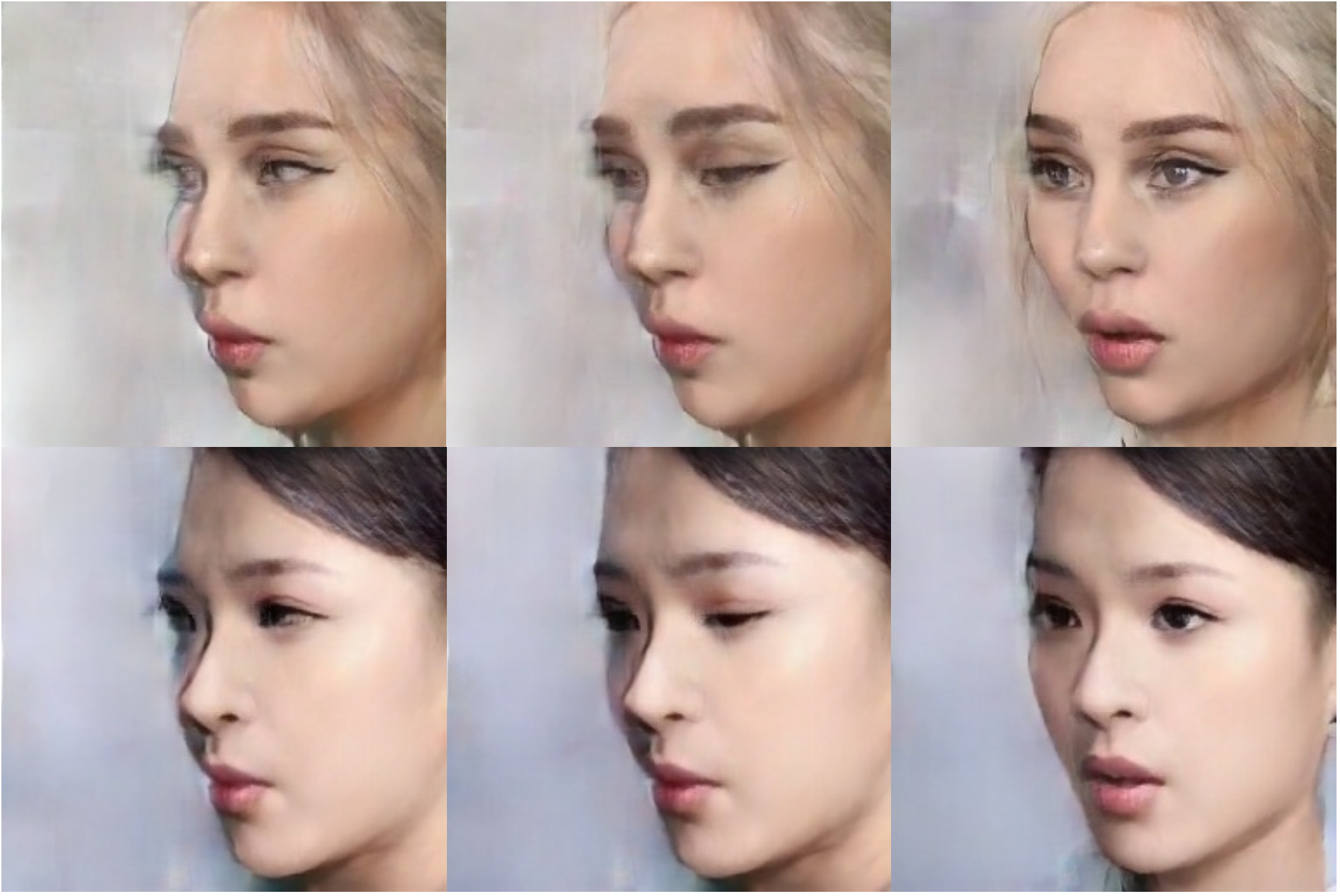}
    }
\caption{ Ablation study of the effectiveness of pose-aware data argumentation on the BeautySelfie dataset. Synthesized results from models with the different setting of pose-aware data argumentation are shown in (a) and (b) respectively.
}
\label{fig:ab2}
\end{figure}

\textbf{Effect of pose-aware data augmentation.}~~As shown in Figure~\ref{fig:ab2}, we compare our model under different settings of pose-aware data augmentation on the BeautySelfie dataset. We observe fewer artifacts and clearer boundary lines on the synthesized side-view faces using the model trained with developed pose-aware data augmentation strategy. This well validates the effectiveness of the proposed data augmentation strategy to improve the capability of our model in transferring poses with large variations.





\section{Conclusion}
In this work, we propose a fine-grained ID-preserving  landmark synthesis approach for deep face reenactment, addressing the problem of used coarse landmark lacking details and landmark ID alternation for the first time. Also, we design several novel losses as training objectives to deal with result blur and artifact which are ubiquitous in the previous studies. Finally, we develop large pose data augmentation strategy coupled with face alignment, to solve the side-view face reenactment problem due to large pose changes. The qualitative and quantitative experimental results demonstrate the superiority of our approach over the state-of-the-art works.


\begin{quote}
\begin{small}

\end{small}
\end{quote}

\end{document}


\maketitle
This is a supplementary material for the main paper. The first section describe the details about our new proposed Beautyselfie dataset. The second section presents additional ablation study on the effectiveness of the proposed loss functions. The last section demonstrates additional qualitative examples to verify the superiority of the extended landmarks.

\begin{table*}[t]
\begin{center}
\begin{tabular}{l c c c c c c c c }
\toprule
Dataset & Expression & Subject & Image & Pose & Resolution & Pose List & Keypoint \\
\hline 
\hline
CK+~ & $8$ & $123$ & $327$ & $1$ & $490X640$ & $F$ & $-$ \\
JAFFE~ & $7$ & $10$ & $213$ & $1$ & $256X256$ & $F$ & $-$ \\
FER2013~ & $7$ & $-$ & $3.5W$ & $-$ & $48X48$ & $-$ & $-$ \\
KDEF~ & $7$ & $140$ & $4.9K$ & $5$ & $562X762$ & $FL,FR,HL,F,HR$ & $-$ \\
F2ED~ & $54$ & $119$ & $21W$ & $4$ & $490X640$ & $F$ & $68points$ \\
\hline
Ours & $7$ & $2.3W$ & $30W$ & $3$ & $720X1280$ & $HL,F,HR$ & $87points$\\
\bottomrule
\end{tabular}
\end{center}
\caption{Comparison between BeautySelfie with existing facial datasets. In the pose list, the full name for each short hand is as follows: F for front, FL for full left, HL for Half left, FR for full right, HR for half right, BV for bird view. }
\label{tab:q1}
\end{table*}

\begin{table*}[t]
\begin{center}
\begin{tabular}{l c c c c c c c}
\toprule
Expression index~\ & $1$ & $2$ & $3$ & $4$ & $5$ & $6$ & $7$ \\
\hline
Expression type~\ & $angry$ & $disgust$ & $fear$ & $happy$ & $sad$ & $surprised$ & $neutral$ \\
\hline
Image number per expression~\ & $42353$ & $32163$ & $18356$ & $56781$ & $23687$ & $26787$ & $99873$ \\
\bottomrule
\end{tabular}
\end{center}
\caption{BeautySelfie dataset is divided into 7 categories, and the chart shows the categories and image number of each facial expression.}
\label{tab:q2}
\vspace{-1em}
\end{table*}

\begin{figure}[t]
\begin{center}
\includegraphics[width=0.9\linewidth]{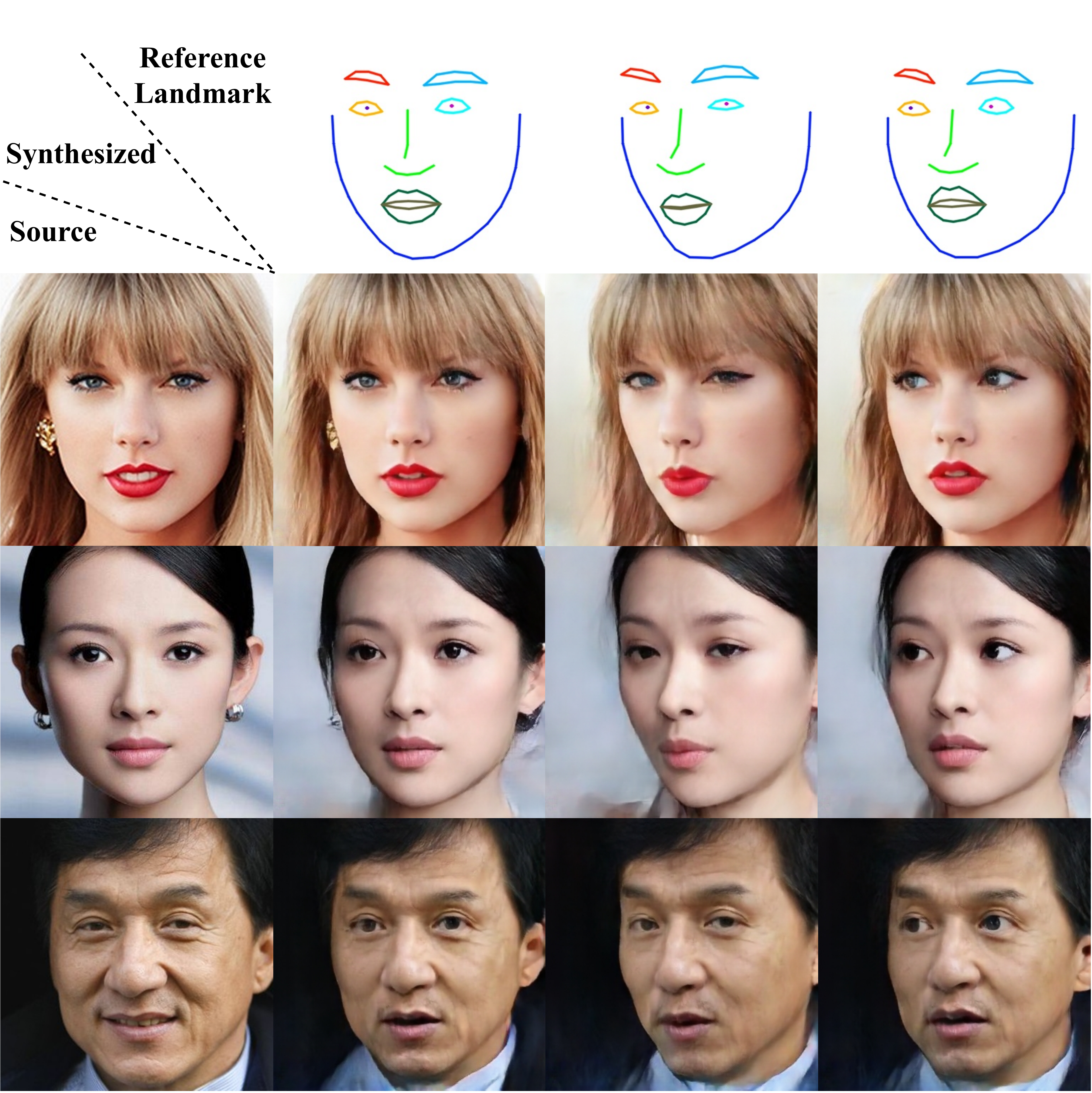}
\end{center}
   \caption{\textbf{ Face reenactment results on our Beautyselfie dataset.} Given fine-grained exemplar landmarks and input faces, our model can synthesize reenactment faces with more dynamic details, including eyeball movement, lip changes and pose variations as clearly demonstrated in the figure.
   }
\label{fig:main}
\end{figure}
\section{Beautyselfie Dataset}
\label{sec:dataset}

\begin{figure*}[ftb]
\begin{minipage}[t]{0.325\textwidth}
\centering
\includegraphics[width=\linewidth]{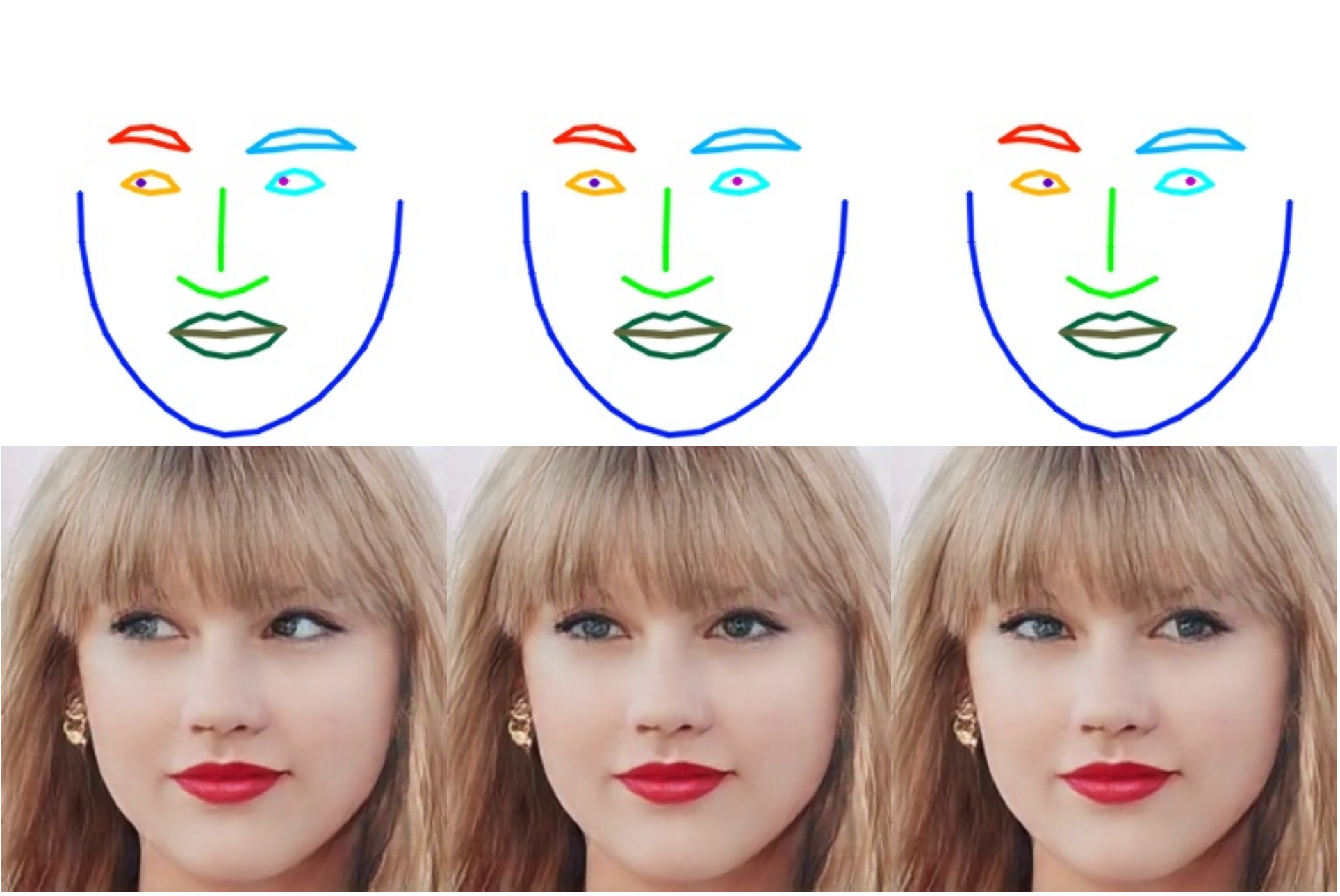}
\subcaption{eyeball position (left $\to$ right)}\medskip
\end{minipage}
\hfill
\begin{minipage}[t]{0.325\textwidth}
\centering
\includegraphics[width=\linewidth]{nose_width.pdf}
\subcaption{nose width (narrow $\to$ wide)}\medskip
\end{minipage}
\hfill
\begin{minipage}[t]{0.325\textwidth}
\centering
\includegraphics[width=\linewidth]{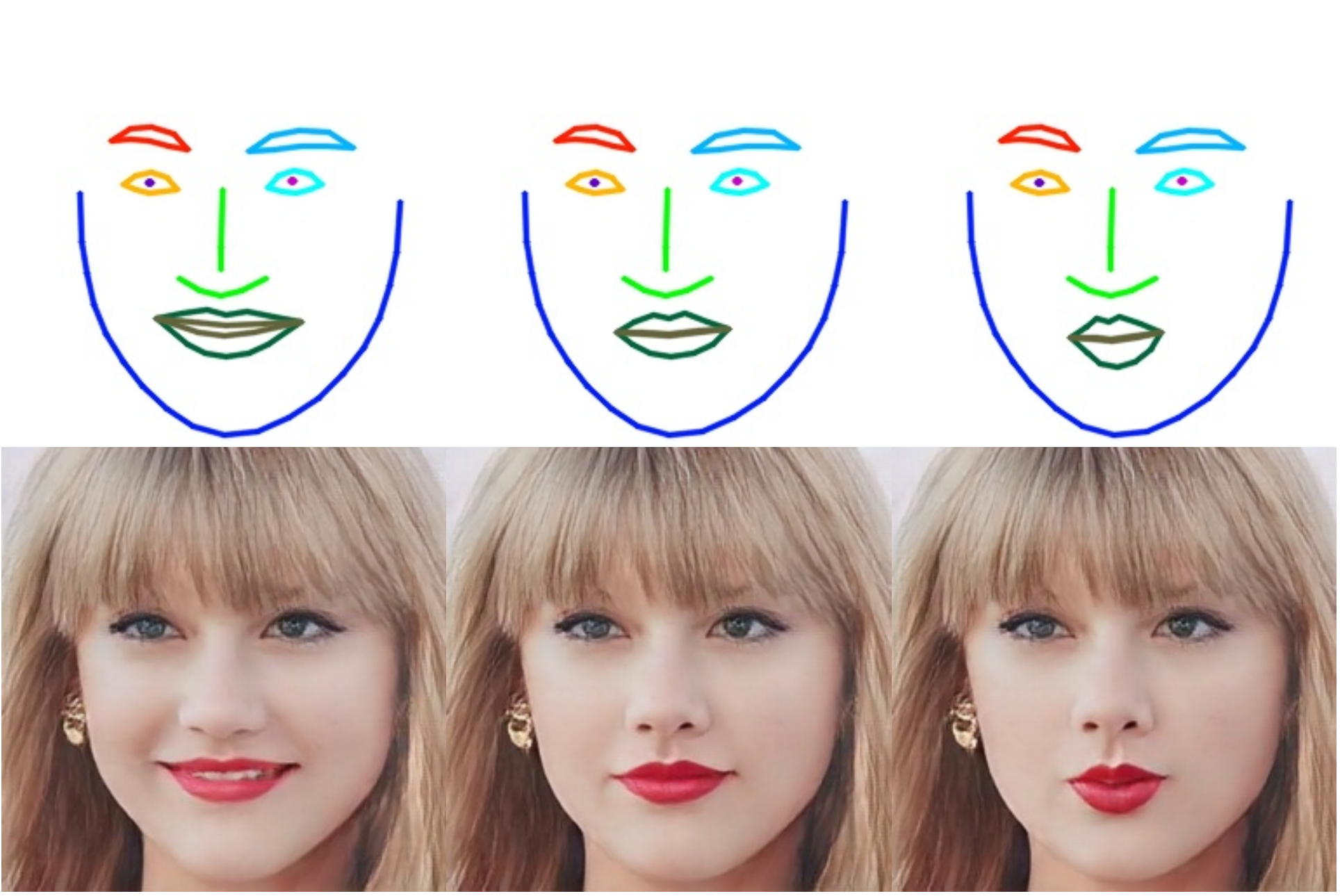}
\subcaption{lip depth (thin $\to$ thick)}\medskip
\end{minipage}
\caption{\textbf{ Face manipulation results on our Beautyselfie dataset.} Manipulating different facial attributes by adjusting eyeball position (a), nose width (b), lip depth (c) of the fine-grained landmarks for conditional face reenactment. }
\label{fig:landmark}
\end{figure*}

To be best of our knowledge, we create the largest high resolution asian facial dataset, namely BeautySelfie, to the community. This data set is collected from $23,000$ subjects, with ages ranging from 18 to 60 and balanced gender ratio. Each subject is firstly trained, to teach them how to present different poses and expressions according to our task requirement. The whole dataset collection lasts a couple of months. Note that the current public facial expression datasets including CK+~\cite{lucey2010extended}, JAFFE~\cite{lyons1998coding}, FER2-13~\cite{carrier2013challenges}, KDEF~\cite{lundqvist1998karolinska}, F2ED~\cite{wang2019fine} are usually collected in the wild or controlled environment, but our BeautySelfie dataset is collected from common life scenes, so it has better generalization ability. Specifically, our BeautySelfie dataset contains $23,000$ video clips, with each clip corresponing to one specific subject and more than 100 images.

We extract $300,000$ representative frames evenly from these video clips, covering all $23,000$ identities and a variety of facial emotions. Table~\ref{tab:q1} shows a comprehensive comparison between our BeautySefie and state-of-the-art facial datasets including CK+, JAFFE, FER2-13, KDEF, F2ED. As shown in Table~\ref{tab:q1}, our dataset contains $23,000$ subjects, while other datasets only contain less than two hundred subjects with low image resolutions. The image numbers of CK+, KDEF, FER2013, JAFFE and F2ED are all much smaller than ours. In particular, our dataset has three times as much data as the largest F2ED. For the image resolution, noting that high resolution is very important for fine-grained face reenactment, as it reserves rich detailed facial part information for high-fidelity face reenactment. Our BeautySelfie dataset has the highest image resolution of $720\times1080$ pixels. To sum up, our dataset has the largest number of subjects and images with highest image quality (resolution).

To show the distribution of each expression type in the dataset, we further give the number of images belonging to each expression type. Table~\ref{tab:q2} shows the distribution. It can be seen from the table that among the seven expressions in our dataset, the fear expression has the smallest number of images ($18,356$), while the neutral expression has the largest number of images, which is close to $100,000$. 
\section{Quantitative Analysis}
\label{exp}

We perform extended ablation study to demonstrate the effectiveness of our proposed loss functions in this section.

As introduced in the main paper, we attempt to improve the quality of synthesized facial images by combining different training objectives, \ie \textit{identity loss}, \textit{mask loss} and \textit{boundary loss}. As shown in Table~\ref{tab:ab1}, stable improvements on both \textbf{SSIM} and \textbf{CSIM} metrics showcase the effectiveness of these loss functions, which enables our model to synthesize images with fewer artifacts while retaining more appearance information of the source faces.

\begin{table}
\begin{center}
\begin{tabular}{l c c c}
\toprule
Method & SSIM$\uparrow$ & CSIM$\uparrow$ \\
\hline 
\multicolumn{3}{c}{BeautySelfie} \\
\hline
vanilla & $0.75$ & $0.61$ \\
identity & $0.75$ & $0.63$ \\
identity + mask & $0.77$ & $0.65$ \\
identity + mask + boundary & $\mathbf{0.79}$ & $\mathbf{0.68}$\\
\bottomrule
\end{tabular}
\end{center}
\caption{ Quantitative results of our approach over different loss settings on the BeautySelfie dataset. }
\label{tab:ab1}
\end{table}

\section{Qualitative Analysis}
\label{ade}

In this section, we present additional examples for qualitative analysis of the functionality of fine-grained landmarks on our Beautyselfie dataset. Fig.\ref{fig:landmark} and Fig.\ref{fig:main} show the image samples of our results. In addition, we also provide synthesized videos to demonstrate the potential of our model in synthesizing video sequences.



\subsection{Additional Video Examples}
Our model is also capable of synthesizing continuous video sequences without constraint of explict temporal consistency. For more details, please refer to the video files attached in the supplementary material.


\begin{quote}
\begin{small}
\bibliography{sample-base}
\end{small}
\end{quote}